\documentclass{article} 
\usepackage{iclr2026_conference,times}

\usepackage{amsmath,amsfonts,bm}

\def\eqref#1{equation~\ref{#1}}

\def\1{\bm{1}}

\DeclareMathAlphabet{\mathsfit}{\encodingdefault}{\sfdefault}{m}{sl}
\SetMathAlphabet{\mathsfit}{bold}{\encodingdefault}{\sfdefault}{bx}{n}

\usepackage{hyperref}
\usepackage{url}
\usepackage{booktabs}
\usepackage{graphicx}
\usepackage{amsmath}
\usepackage{diagbox}
\usepackage{amssymb}
\usepackage{pifont}
\usepackage{float}
\usepackage{tikz}
\usepackage{tcolorbox}
\usepackage{url}
\usepackage[dvipsnames]{xcolor}
\definecolor{newcolor}{rgb}{.8,.349,.1}
\usepackage{enumitem}
\usepackage{colortbl}
\usepackage{float}
\usepackage[ruled]{algorithm2e}
\usepackage{subcaption}
\usepackage{multirow}
\usepackage{enumitem}
\usepackage[T1]{fontenc}
\usepackage{caption}   
\tcbuselibrary{breakable}

\DeclareRobustCommand{\legendbg}[2]{%
\unskip
  \begingroup\setlength{\fboxsep}{0pt}
  \colorbox{#1}{\strut #2}%
  \endgroup
}

\DeclareRobustCommand{\legendbgp}[2]{%
\unskip
  \begingroup\setlength{\fboxsep}{0pt}
  (\colorbox{#1}{\strut #2})
  \endgroup
\ignorespaces
}

\definecolor{mygray}{gray}{.9}
\newtcolorbox[list inside=prompt,auto counter,number within=section]{prompt}[1][]{
    colbacktitle=black!60,
    coltitle=white,
    fontupper=\footnotesize,
    boxsep=5pt,
    left=0pt,
    right=0pt,
    top=0pt,
    bottom=0pt,
    boxrule=1pt,
    title={#1},
    breakable,
    #1, 
}

\newcommand{\agentsymbR}{2.9pt}      
\newcommand{\agentsymbDelta}{1.1pt}  
\newcommand{\agentsymbLW}{0.6pt}     

\newcommand{\nonagentic}{%
  \tikz[baseline=-0.6ex]\fill (0,0) circle (1.6pt);}

\newcommand{\singleagent}{%
  \tikz[baseline=-0.6ex]\draw[line width=\agentsymbLW] (0,0) circle (\agentsymbR);}

\newcommand{\multiagent}{%
  \tikz[baseline=-0.6ex]{
    \draw[line width=\agentsymbLW] (0,0) circle (\agentsymbR);
    \draw[line width=\agentsymbLW] (0,0) circle (\agentsymbR+\agentsymbDelta);
  }}

\newcommand{\eg}{e.g.}

\newcommand{\revise}[1]{{#1}}
\title{\methodname{}: A Trainable Hierarchical Automaton System for Multi-Agent Visual Reasoning}

\author{\textbf{Zhixi Cai,  Fucai Ke, Kevin Leo, Sukai Huang, Maria Garcia de la Banda, Peter J. Stuckey,}\\ \textbf{Hamid Rezatofighi}\\
Monash University, Australia\\
\texttt{\{zhixi.cai,fucai.ke1,kevin.leo,sukai.huang,maria.garciadelabanda,}\\
\texttt{peter.stuckey,hamid.rezatofighi\}@monash.edu}
}

\newcommand{\methodname}{MATA}
\newcommand{\methodfull}{Multi-Agent hierarchical Trainable Automaton}
\newcommand{\datasetname}{\methodname-SFT-90K}

\iclrfinalcopy 
\begin{document}

\maketitle

\begin{figure}[ht]
\centering
\includegraphics[width=\textwidth]{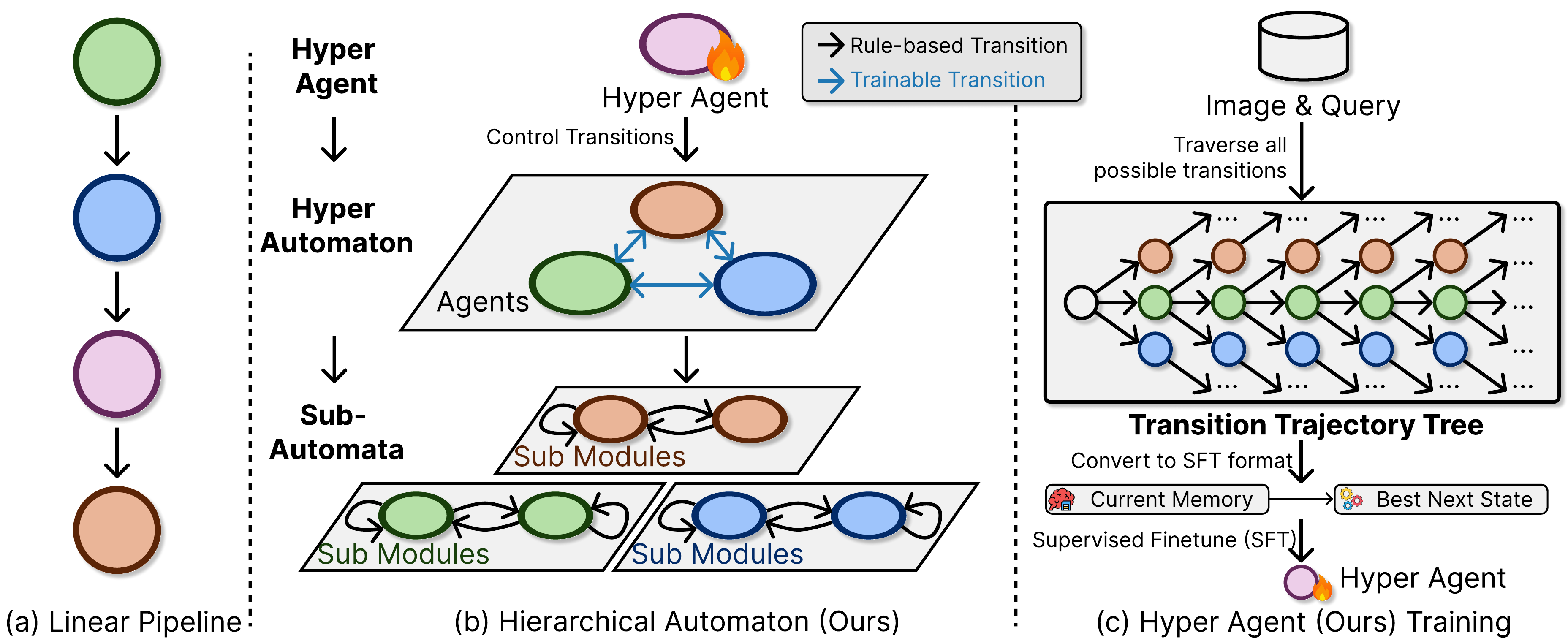}
\caption{\textbf{Overview of \methodname{}.} \textbf{(a)} Linear pipelines (previous methods) execute modules in a fixed, manually designed order. 
\textbf{(b)} \methodname{} organizes agents as states in a hyper automaton. A trainable hyper agent learns high-level transitions between agents (blue arrows), enabling collaboration and competition, while each agent runs a small rule-based sub-automaton for reliable micro-control (black arrows). 
\textbf{(c)} To train the hyper agent, we expand a transition-trajectory tree per image-query, score the leaves using task metrics, and convert each node’s snapshot into a supervised pair \emph{current memory $\rightarrow$ best next state} for supervised finetuning (SFT), forming \datasetname{}.}
\label{fig:teaser}
\end{figure}

\begin{abstract}
Recent vision-language models have strong perceptual ability but their implicit reasoning is hard to explain and easily generates hallucinations on complex queries. Compositional methods improve interpretability, but most rely on a single agent or hand-crafted pipeline and cannot decide when to collaborate across complementary agents or compete among overlapping ones. We introduce \methodname{} (\methodfull{}), a multi-agent system presented as a hierarchical finite-state automaton for visual reasoning whose top-level transitions are chosen by a trainable hyper agent. Each agent corresponds to a state in the hyper automaton, and runs a small rule-based sub-automaton for reliable micro-control. All agents read and write a shared memory, yielding transparent execution history. To supervise the hyper agent's transition policy, we build transition-trajectory trees and transform to memory-to-next-state pairs, forming the \datasetname{} dataset for supervised finetuning (SFT). The finetuned LLM as the transition policy understands the query and the capacity of agents, and it can efficiently choose the optimal agent to solve the task. Across multiple visual reasoning benchmarks, \methodname{} achieves the state-of-the-art results compared with monolithic and compositional baselines. The code and dataset are available at \url{https://github.com/ControlNet/MATA}.
\end{abstract}

\clearpage

\section{Introduction}

Visual reasoning is the cognitive process of interpreting and analyzing relationships among entities in a visual scene to support decision‑making and problem‑solving~\citep{ke_explain_2025}. Although recent Vision-Language Models (VLMs)~\citep{liu_visual_2023, chen_internvl_2024, bai_qwen25-vl_2025} demonstrate strong perceptual ability, their implicit reasoning is difficult to audit and often causes hallucinations on complex queries involving spatial relations, spatial attributes, and counting. Compositional approaches~\citep{suris_vipergpt_2023, you_idealgpt_2023, ke_hydra_2024, cai_naver_2025} improve interpretability by decomposing a task into planning, perception, and reasoning stages, typically employing Large Language Models (LLMs)~\citep{gemini-team_gemini_2023, openai_gpt-4o_2024, deepseek-ai_deepseek-r1_2025} as planners or code generators and Vision Foundation Models (VFMs)~\citep{radford_learning_2021, liu_grounding_2023, xiao_florence-2_2024, yang_depth_2024} as perceptual tools. Despite these improvements, non-agentic compositional methods~\citep{suris_vipergpt_2023, lu_chameleon_2023} struggle in practice: they are limited to a single-turn reasoning, thus lacking the ability to incrementally reason in a closed-loop. Due to these limitations, agentic methods~\citep{you_idealgpt_2023, ke_hydra_2024, gao_clova_2024, zhong_viotgpt_2025} treat visual reasoning as a multi-step feedback loop in which agents actively take actions based on the current state~\citep{ke_explain_2025}.

However, most agentic systems still employ a single agent, which is often insufficient for complex reasoning~\citep{wang_towards_2025} as different skills are required for different parts of a problem. 
Further, in prior multi-agent methods~\citep{hong_metagpt_2023, li_agent-oriented_2024, nguyen_ma-rag_2025, zhang_agentorchestra_2025} (developed for other domains), \emph{collaborative} agents are assigned disjoint roles for different subtasks and are organized into hard-coded pipelines. While this is simple and interpretable, it prevents error and hallucination handling, and tends to propagate upstream mistakes through the pipeline~\citep{gao_clova_2024, ke_dwim_2025}. In contrast, a \emph{competition} mechanism where functionally overlapping agents for the same subtask work together is under-explored in previous work.
In this paper, we explore compositional multi‑agent visual reasoning in an environment where collaborative and competitive agents exist.

Motivated by the requirements above, we cast this decision problem as a finite-state automaton where the transition function picks a discrete next state and the lifecycle is naturally expressed by explicit states and transitions. This provides explainability, verifiable control flow, and modularity that yield greater versatility, reliability, and performance.  A recent work~\citep{cai_naver_2025} also used an automaton, but its hand‑written rule-based transitions are inflexible and difficult to manually define as states and transitions grow~\citep{wang_learning_2025, yue_synergistic_2025, dang_multi-agent_2025, wan_rema_2025}. When new agents are added, their transitions need to be manually defined. Designing rules to select among functionally overlapping (competitive) agents is hard since the criteria are ambiguous and task‑dependent, and human priors about which agents fit which tasks and queries are uncertain. We therefore design a trainable hyper agent to learn a transition policy that selects the next state. Notably, not every transition needs learning: within an agent, micro‑steps (e.g., LLM/VLM prompting, verifier checks, tool I/O) follow clear procedures that are easy to define. As the number of agents grows, the main difficulty is cross‑agent transition rather than agent's inside control. This motivates a hierarchical automaton in which each top‑level state is an agent with a small rule‑based sub‑automaton, and a trainable hyper agent provides the transition function that  observes the shared memory and selects the next agent. All agents read and write to a shared memory that records variables, tool outputs, code history, and verifier feedback, recording an explainable process. This replaces an inflexible rule-based transition policy with a data‑driven, error‑aware, and dynamic policy that can redirect to alternative solutions when needed. This design focuses on learning the ambiguous selection between competitive agents, while preserving reliable execution inside agents. 

We introduce these ideas in \methodname{} (\methodfull{}), a hierarchical automaton for visual reasoning. \methodname{} contains a specialized agent for fast, System~1-style perception (e.g., object detection, simple question answers); a slow, System~2-style step-wise reasoner that generates and executes Python programs for multi-step inference; and a one-shot workflow reasoner that solves queries without iteration.

To supervise the hyper agent, we need labeled transition decisions. We therefore run the system for each image-query pair, expand a transition trajectory tree~\citep{kearns_sparse_2002} and log the state history, prompts, intermediate artifacts (detections, captions, code), feedback, and performance results. The leaves are scored by the appropriate task performance, and each decision is labeled with the child that leads to the highest‑scoring subtree. This generates memory‑to‑next‑state pairs (\datasetname{}) for LLM supervised finetuning (SFT), as shown in \autoref{fig:teaser} (c).

The contributions of our paper are:

\begin{itemize}[leftmargin=0.35cm]
    \item A hierarchical deterministic finite-state automaton-based system, \methodname{}, that unifies neuro-symbolic framework with collaborative and competitive multi-agent design for visual reasoning.
    \item Proposing (i) a learnable mechanism that trains a hyper agent as the transition policy of the hyper automaton over collaborative and competitive agents; (ii) a transition-trajectory data generation pipeline and the dataset, \datasetname{}, for supervised finetuning (SFT) of the hyper agent.
    \item Comprehensive experiments across visual-reasoning benchmarks, with extensive ablations and analysis.
\end{itemize}

\section{Related Works}

Monolithic vision-language models (VLM) map images and text directly to answers with a single forward pass~\citep{xiao_florence-2_2024, liu_grounding_2023, li_otter_2023, li_blip-2_2023, wu_next-gpt_2023, stanic_towards_2024, zhu_minigpt-4_2023}. While these models have strong perceptual capabilities, their implicit reasoning processes are hard to explain and often degrade on queries requiring spatial relations, counting, or multi-step reasoning~\citep{jahangard_jrdb-social_2024, jahangard_jrdb-reasoning_2025}. This motivates modular designs that expose intermediate, explainable symbolic processes~\citep{andreas_neural_2016, hsu_whats_2023}. Compositional methods decompose a task into multiple stages~\citep{ke_explain_2025}, often by having an LLM generate grounded actions (\eg, programs or JSON) executed by tools~\citep{gupta_visual_2023, suris_vipergpt_2023, shen_hugginggpt_2023, lu_chameleon_2023}. These pipelines improve interpretability and enable external tools use, but usually operate in a single forward pass with a fixed manually designed pipeline. They thus lack a flexible mechanism to engage in multi-step reasoning from feedback.

Recent works~\citep{you_idealgpt_2023, ke_hydra_2024, gao_clova_2024, zhong_viotgpt_2025} explore agentic systems where an LLM/VLM reasons in multiple steps and calls tools~\citep{ke_explain_2025}. However, most agentic approaches in visual reasoning remain single-agent. In broader domains, multi-agent frameworks assign disjoint roles and connect them with hand-crafted collaboration patterns~\citep{hong_metagpt_2023, li_agent-oriented_2024, nguyen_ma-rag_2025, zhang_agentorchestra_2025}, achieving better performance in reasoning. However, this idea is still under-explored for visual reasoning. Notably, noise from perception and LLM/VLM hallucinations can accumulate across steps~\citep{ke_dwim_2025} from the collaborating pipelines, and most designs overlook competition between functionally overlapping agents~\citep{wang_towards_2025}. This lack of a learned transition policy limits flexibility and robustness on complex and diverse queries.

Finite-state automata as abstractions provide explicit control flow and interpretability. NAVER introduces probabilistic logic inside an automaton and equips modules with self-correction~\citep{cai_naver_2025}, but relies on a hand-crafted transition policy that is hard to manually define as states grow. HYDRA introduces an agent that includes a planner, an RL controller, and a code-executing reasoner~\citep{ke_hydra_2024}. While data-driven, it still focuses on instruction-level planning without a learned, high-level policy for switching across qualitatively different agents on demand.
\revise{By contrast, we propose \methodname{} that explicitly learns the inter-agent transition function over a hyper-automaton whose states are agents, while keeping intra-agent micro-steps rule-based. This learned transition function enables collaboration and competition among overlapping experts and transfers across different domains and tasks (\autoref{para:generalizability}), which previous visual reasoning methods with hand-written transitions or single-agent controllers do not address.} States are agents; each agent runs a small, rule-based sub-automaton for reliable micro-control, while a trainable hyper agent learns cross-agent transitions over a shared memory. This hierarchical view retains the interpretability of explicit state machines, avoids hand-coded transitions, and supports both collaboration and competition. Unlike prior work~\citep{ke_hydra_2024, cai_naver_2025}, our controller is supervised-trained from transition-trajectory data to transit between agents and to report a final result only when it is certain of the answer, directly addressing the gap identified above.

\begin{figure}[t]
\centering
\includegraphics[width=\textwidth]{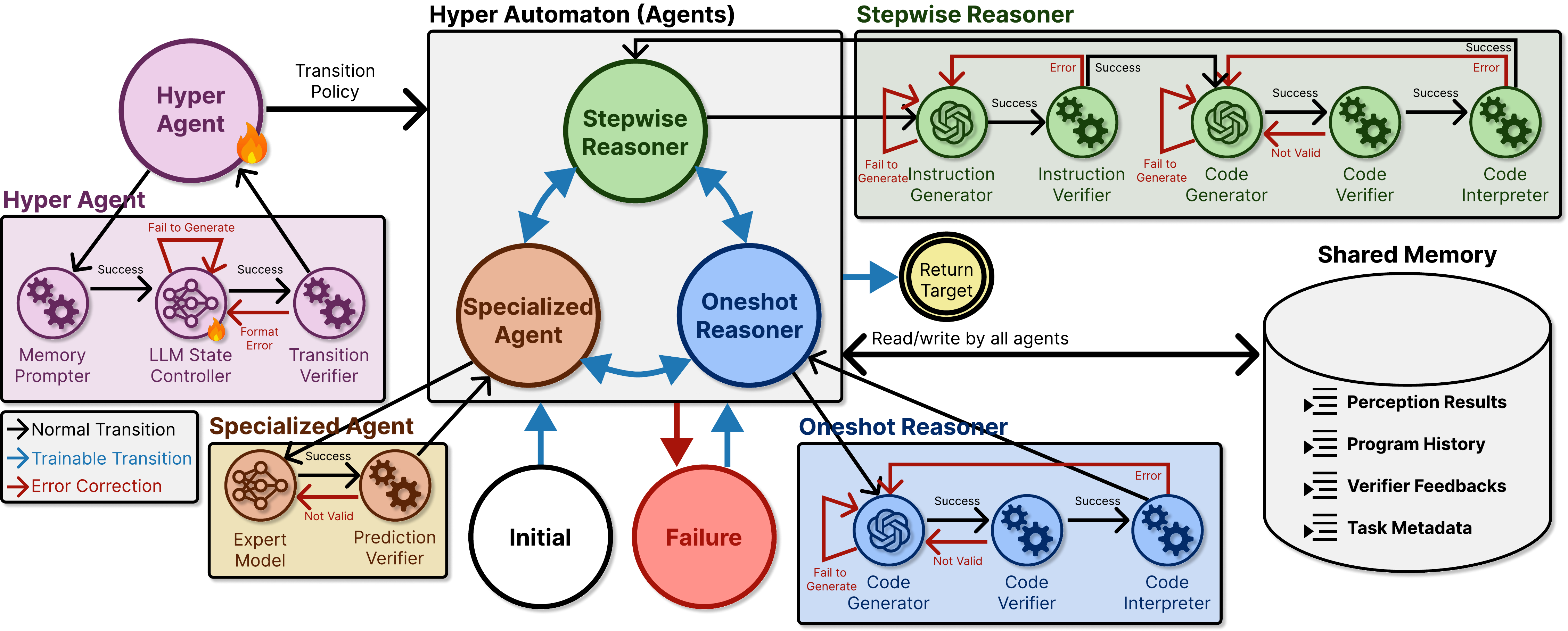}
\caption{\textbf{Pipeline of \methodname{}.} A trainable \emph{hyper agent} reads a snapshot of the shared memory, predicts the next state with an \emph{LLM State Controller}. Its decision (blue arrows) routes control among agent states in the \emph{hyper automaton}:  Oneshot Reasoner, Stepwise Reasoner, and Specialized Agent. Each agent runs a rule-based sub-automaton that iterates until return to the hyper automaton. All agents read/write an append-only \emph{Shared Memory}, enabling the hyper agent to access the current context for choosing the optimal next state. Lifecycle states \textsc{Initial} and \textsc{Failure} are shown outside the agents (see \autoref{sec:hyper_automaton} for details).}
\label{fig:pipeline}
\end{figure}

\section{Methodology}

We explore multi-agent visual reasoning by learning a high-level transition function over agents within a hierarchical automaton, enabling data-driven \emph{collaboration} and \emph{competition} among overlapping skills and replacing inflexible hand-written pipelines.

\subsection{Overview}
\label{subsec:problem_setting}

A visual reasoning instance is an image-query pair $(v,q)$ mapped to an output $y$~\citep{ke_explain_2025}. \methodname{} organizes inference as a \emph{hierarchical automaton} operated by a trainable \emph{hyper‑agent}. Informally, the hyper automaton $\mathcal{M}_\theta$ is a top-level automaton whose states include a set of sub-agents, with each sub-agent running a small rule-based sub-automaton, and the trainable hyper agent controlling the learned transition $\delta_\theta$. Formally, it can be described as a Mealy machine~\citep{mealy_method_1955}: $\mathcal{M}_\theta = (S, S_0, \Sigma, \Lambda, \delta_\theta, \Gamma)$ where $S$ denotes the set of states (containing both agent states for task execution and lifecycle states for process coordination), $S_0$ the initial state where reasoning begins, $\Sigma$ the inputs drawn from shared-memory snapshots (storing intermediate results from agents), $\Lambda$ the answer space of visual reasoning queries (e.g., discrete labels, bounding box coordinates, or free-text responses), $\delta_\theta$ the learned transition function that determines the next state based on the current state and memory inputs, and $\Gamma$  the output function that generates the final answer $\hat y$ once the automaton reaches a terminal state. Detailed breakdowns of the states, transition mechanics, and output generation process are provided in the subsequent sections (\autoref{fig:pipeline}).

\subsection{Hyper Automaton} \label{sec:hyper_automaton}

\paragraph{States.} The finite state set is the union of \emph{agent states} and \emph{lifecycle states}: $S = S_\text{agent} \cup S_\text{life}$, where $S_\text{agent} = \{ \textsc{Oneshot}, \textsc{Stepwise}, \textsc{Specialized}\}$, $S_\text{life} = \{\textsc{Initial}, \textsc{Final}, \textsc{Failure}\}$ and the initial state $S_0 = \textsc{Initial}$. Agent states invoke concrete skills; lifecycle states orchestrate the progression and termination of the reasoning episode (e.g., starting the task, handling uncertainty, concluding with an answer). Details of the states are shown in \autoref{tab:states}.

\begin{table}[h]
\centering
\caption{\textbf{States of the hyper automaton.} The table specifies the description and the triggering condition for each state. $\delta_\theta$: transition function of hyper automaton.}
\label{tab:states}
\scalebox{0.9}{\begin{tabular}{l|p{0.75\linewidth}|p{0.124\linewidth}}
\toprule[0.4mm]
\rowcolor{mygray}\textbf{State} & \textbf{Description} & \textbf{Selected by}\\
\midrule
\textsc{Initial} & The unique state where reasoning begins.& Initial state \\ \hline
\textsc{Oneshot} & A workflow agent that executes a single-pass program generation and execution workflow for solvable queries, equipped with a lightweight verifier. & \multirow{9}{*}{{\Large$\delta_\theta$}} \\
\textsc{Stepwise} & A stepwise reasoner that produces step-wise Python programs for complex queries; code is verified and
executed in a sandboxed environment to ensure correctness. & \\
\textsc{Specialized} & An expert agent that performs fast perception tasks; built-in verifiers validate outputs and adapt parameters.& \\
\textsc{Final} & A terminal state in which sufficient evidence has accumulated; the output function $\Gamma$ is invoked when in this state to produce the final answer $\hat{y}$. & \\ \hline
\textsc{Failure} & A state triggered by unrecoverable errors or exceeding iteration limit.& Error occurs \\
\bottomrule[0.4mm]
\end{tabular}}
\vspace{3mm}
\end{table}

Agents in our system are intentionally both \emph{collaborative} and \emph{competitive}. When control transitions from one agent to another, the successor agent reads the shared memory containing the prior history and feedback, and builds on that context; this is \emph{collaboration}. At the same time, multiple agents may attempt the same task; if one agent stalls or fails, another can take over and complete it; this is \emph{competition}. \revise{The learned transition policy $\delta_\theta$ selects among them based on context (e.g., \textsc{Oneshot} vs.\ \textsc{Stepwise} for moderately compositional VQA; \textsc{Specialized} vs.\ \textsc{Oneshot} for grounding with simple perception). This overlap is intentional, as the three agents represent a spectrum: \emph{perception (system 1)}, \emph{one-shot reasoning (fast thinking)}, and \emph{stepwise reasoning (slow thinking)}. Although all agents can answer all queries, each agent has different advantages and disadvantages, enabling hyper agent to choose the optimal transition and re-route on failure.} The implementation details of agents are shown in the supplementary material (\autoref{sec:implementation_agents}).

\paragraph{Shared Memory.} All agents read from and write to a structured \emph{shared memory} $m_t$ at the $t$-th step that accumulates intermediate variables, perception results, program history, verifier feedback, and task metadata. We keep the formalism minimal: when an agent runs for one cycle, it appends its new memory $\Delta m_t$, and $m_{t+1}=m_t\cup\Delta m_t$. Memory is append‑only so the full reasoning process is auditable and visible to the hyper agent.

\paragraph{Execution Step.} At step $t$ the system is in $(s_t,m_t)$. The hyper‑agent observes the memory $m_t$ and selects the next state $s_{t+1}$ via the learned transition function $\delta_\theta$:
\begin{equation}
s_{t+1}=\delta_{\theta}(s_t,m_t),\ \ \ s_{t+1}\in S.
\label{eq:transition}
\end{equation}
If $s_{t+1}\in S_{\mathrm{agent}}$, the corresponding agent executes its rule‑based sub‑automaton until returning to the hyper automaton and updating the memory;
if $s_{t+1} = \textsc{Final}$ or $t > T$ where $T$ is the max step limit, the episode terminates.

\paragraph{Output.} 
The answer space $\Lambda$ contains the required output $\hat{y}$ for visual reasoning. For example, $\Lambda = \{y \mid y \text{ is text for VQA, bounding box for VG, etc}\}$. The output function $\Gamma$ extracts the output from the memory $m_t$ at \textsc{Final} state: $\hat{y}=\Gamma(\textsc{Final},m_t)$.

\subsection{Trainable Transition Function (Hyper Agent)}
\label{subsec:transition}

The transition function $\delta_\theta$ in \autoref{eq:transition} is implemented by a trainable LLM-based \emph{hyper agent} $\mathcal{F}_\theta$. This agent acts as the state-transition controller, selecting the next state $s_{t+1}$ from a limited set of available candidate states. Since the LLM requires textual input, we derive a prompt $x_t$ from the shared memory $m_t$. The template for constructing $x_t$ from $m_t$ is shown below:

\begin{prompt}[title={Prompt \thetcbcounter: LLM State Controller in Hyper Agent}] \label{prompt:hyper_agent}
You are an AI assistant to control the state of a multi-step visual reasoning system. Your task is to decide the next state the system should transition to based on the current state and history. \\
<TaskDescription>\{task\_title\}\{task\_description\}</TaskDescription>\\
<Query>\{query\}</Query>\\
<Feedback>\{feedback\}</Feedback>\\
<Code>\{code\}</Code>\\
<Variables>\{variables\}</Variables>\\
<StateHistory>\{state\_history\}</StateHistory>\\
<State>\{state\}</State>\\
<CurrentStep>\{current\_step\}</CurrentStep>\\
Based on the information above, determine the next state the system should transition to. Choose from the following states:\\
<StateCandidates>\{next\_state\_candidates\}</StateCandidates>\\
Return the name wrapped in <NextState> tags.
\end{prompt}

Our hyper agent $\mathcal{F}_\theta$ maps the prompt $x_t$ to a distribution over the available states, from which $s_{t+1}$ is selected, either through greedy decoding or stochastic sampling.

The parameter $\theta$ of the hyper agent is supervised finetuned (SFT) on our collected transition trajectory dataset $\mathcal{D}$ (\autoref{sec:dataset_generation}). Each training example provides a textual memory $x_t$ as prompt and a target next state chosen by scanning branches in the trajectory tree that lead to successful and higher final scores:
\begin{equation}
\theta \leftarrow \arg\min_\theta\mathcal{L}_{\mathrm{SFT}}(\theta;\mathcal{D})
\end{equation}
This objective guides the hyper agent on how to switch between sub-agents, and finalize the output.

\subsection{Dataset Generation}
\label{sec:dataset_generation}

Learning the transition policy of the hyper automaton requires examples of how agent states interact during visual reasoning. We therefore build a dataset of transition trajectories. We regard the set of possible transition trajectories from an initial state as a trajectory tree $\mathcal{T}(v,q)$~\citep{kearns_sparse_2002} that records, for each node: the state history, intermediate reasoning outcomes, and final metric scores, as a textual prompt $x_t$ based on \hyperref[prompt:hyper_agent]{prompt 3.1}. We collect this data by running \methodname{} while systematically traversing each next-state option rather than committing to a single path. Unlike end-to-end LLM/VLM training, this procedure explicitly explores the space of possible agent states and yields labeled decisions for our model.

Concretely, we sample images and queries from the training splits of GQA~\citep{hudson_gqa_2019}, OK-VQA~\citep{marino_ok-vqa_2019}, and RefCOCO/RefCOCO+/RefCOCOg~\citep{kazemzadeh_referitgame_2014} and run the hyper automaton $\mathcal{M}_\theta$ step-wise. Rather than limiting to a single route, we expand a bounded trajectory tree to depth $T$: at each node (state) the controller branches over the possible next states $s_{t+1}\in S$, executes the corresponding sub-automaton, and saves a memory checkpoint $m_{t+1}$. When a terminal state is reached (e.g., \textsc{Final}), which by construction corresponds to a \emph{leaf} of the tree $\mathcal{T}$, the output function $\Gamma$ produces a prediction $\hat y$ for the given image-query pair $(v,q)$ with ground truth $y$. We then compute a scalar task score for that leaf: for VG we use $\mathrm{IoU}(\hat y,y)$; for VQA we use $\mathrm{Acc}(\hat y,y)$. During data collection we perform a near-exhaustive expansion of the transition tree to a fixed depth, which is tractable with the current three agents but, we acknowledge, grows rapidly as more agents/states are added.

\textbf{Bottom-up node scoring.}\quad As a result, each leaf node $s \in \mathrm{Leaves}(\mathcal{T})$ is associated with a prediction $\hat y_s$ and ground truth $y$, from which we compute a scalar score. We assign values to all nodes by propagating these scores upward from the leaves:
\begin{equation}
V(s) \triangleq 
\begin{cases}
\mathrm{metric}(\hat y_{s}, y), & s \in \mathrm{Leaves}(\mathcal{T}),\\
\max_{s'\in \mathrm{Child}(s)} V(s'), & \text{otherwise.}
\end{cases}
\end{equation}

To train the LLM state controller, we convert each multi-choice transition into supervised examples. For every decision point at state $s_t$ with corresponding textual prompt $x_t$, we determine the optimal next state $s_t^{\star}$ by selecting the child node that leads to the subtree with the highest propagated value. Formally, for a state $s_t$ with its set of next states $\mathrm{Child}(s_t) \subseteq S$, we choose:

\begin{equation}
s_t^{\star} \in \arg\max_{s \in \mathrm{Child}(s_t)} V(s).
\end{equation}

The chosen state $s_t^{\star}$ becomes the label for the corresponding node prompt $x_t$, and together they form a training example. Repeating this over all decision points produces a dataset of message histories paired with optimal next states, $\mathcal{D}=\{(x_i,\, s_i^{\star})\}_{i=1}^{N}$.
Finally, we reformat the collected examples into instruction-completion pairs suitable for supervised finetuning of LLM. Training on this dataset enables the model to learn how to control the transitions of a hyper automaton. In total, we build the SFT dataset \datasetname{} containing $N=90{,}854$ examples. \revise{We show the data example in \autoref{sec:dataset_example}.}

\subsection{Inference}
\label{sec:inference}

Given an image-query pair $(v,q)$, we initialize the shared memory $m_0$ and enter the initial state $s_0=\textsc{Initial}$. At step $t$, the hyper agent $\mathcal{F}_\theta$ reads the current context $x_t$ and selects the next state $s_{t+1}$ using the learned transition in \autoref{eq:transition}. If $s_{t+1}\in S_{\text{agent}}$, the corresponding sub-agent executes one cycle of its rule-based sub-automaton, appends its intermediate result to memory, and returns to the hyper automaton.
If $s_{t+1} = \textsc{Failure}$, this state indicates that the selected agent $s_t$ reports an unrecoverable error and the system will invoke the hyper agent to choose a new state $s_{t+1}$ while temporarily removing the failed agent $s_t$ from the state candidates to avoid infinite retries. If $s_{t+1} = \textsc{Final}$ or the step $t$ exceeds the limit $T$, the system terminates and returns the final result $\hat{y}$. 

\section{Experiments and Results}

\paragraph{Implementation Details.}
\label{subsec:impl}
We implement \methodname{} in PyTorch~\citep{paszke_pytorch_2019} and conduct all experiments on 4 RTX 4090 GPUs. The system uses interchangeable foundation models; unless otherwise stated we adopt InternVL2.5 (8B)~\citep{chen_expanding_2025} as the VLM, Florence2-L~\citep{xiao_florence-2_2024} for object detection, DepthAnythingV2~\citep{yang_depth_2024} for depth, and a Qwen3 (4B)~\citep{yang_qwen3_2025} LLM for the trainable state controller in the hyper agent. The LLM is supervised finetuned on \datasetname{} using AdamW, cosine decay with 5\% warm-up, global batch size 64, for 8 epochs; decoding is guided at inference to ensure the output format. As \datasetname{} is a dataset collected by running our pipeline on multiple source datasets, ``training on dataset X" means training on the subset of \datasetname{} whose trajectories were generated from the training split of X. We use three SFT configurations for the hyper agent: (i) \legendbg{green!20}{domain-specific}: trained on the training split of the target dataset and evaluated on its test split; (ii) \legendbg{yellow!25}{\revise{domain-transfer}}\footnote{\revise{Our \emph{\revise{domain-transfer}} term is scoped to the hyper agent: it is trained on non-test-dataset transition trajectories, and never sees the optimal trajectories in other datasets.}}: trained on the dataset which is not the target dataset for \revise{evaluation}; and (iii) \legendbg{blue!15}{general}: trained jointly on the whole dataset. We follow the official splits of all the benchmark datasets, reporting accuracy. For fairness, when comparing with compositional baselines we keep the same foundation models, and for monolithic models we use the available public checkpoints with their official code. 
\revise{In the inference, we limit the max step of \methodname{} $T=15$ to avoid infinite running. The prompt template is shown in the \autoref{sec:prompt_template} in supplementary material.}

\paragraph{Evaluation Protocol.}
We evaluate on GQA~\citep{hudson_gqa_2019}, OK-VQA~\citep{marino_ok-vqa_2019}, RefCOCO/RefCOCO+/RefCOCOg~\citep{kazemzadeh_referitgame_2014}, and Ref-Adv~\citep{akula_words_2020} following the previous works~\citep{suris_vipergpt_2023, ke_hydra_2024, cai_naver_2025}, with accuracy as the metric. We compare against the previous compositional methods which are training-required~\citep{khan_self-training_2024, ke_dwim_2025} or training-free~\citep{suris_vipergpt_2023, ke_hydra_2024, cai_naver_2025}, and monolithic methods~\citep{li_blip-2_2023, zhu_minigpt-4_2023, liu_visual_2023, su_pandagpt_2023, han_imagebind-llm_2023, dai_instructblip_2023, li_otter_2023, wang_qwen2-vl_2024, bai_qwen25-vl_2025, chen_expanding_2025, zhu_internvl3_2025, wang_internvl35_2025, openai_gpt-4o_2024, tiong_plug-and-play_2022, yang_empirical_2022, alayrac_flamingo_2022}.

\begin{table}[tb]
\centering
\begin{minipage}[t]{0.49\textwidth}
\centering
\captionof{table}{\textbf{Performance on GQA dataset.}}
\label{table:gqa}
\scalebox{0.8}{
\begin{tabular}{c|c|l|c}
\toprule[0.4mm]
\rowcolor{mygray} \multicolumn{2}{c|}{\textbf{Type}} & \textbf{Method} & \textbf{Acc.} \\ 
\hline\hline
\multirow{13}{*}{\rotatebox[origin=c]{90}{Monolithic}} 
    & \nonagentic{} & BLIP-2~\citep{li_blip-2_2023} & 45.5 \\
    & \nonagentic{} & MiniGPT-4 (13B)~\citep{zhu_minigpt-4_2023} & 30.8\\
    & \nonagentic{} & LLaVA (13B)~\citep{liu_visual_2023} & 41.3\\
    & \nonagentic{} & PandaGPT (13B)~\citep{su_pandagpt_2023} & 41.6\\
    & \nonagentic{} & ImageBind-LLM (7B)~\citep{han_imagebind-llm_2023} & 41.2\\
    & \nonagentic{} & InstructBLIP (13B)~\citep{dai_instructblip_2023} & 49.5\\
    & \nonagentic{} & Otter (7B)~\citep{li_otter_2023} & 50.0\\
    & \nonagentic{} & Qwen2-VL (7B)~\citep{wang_qwen2-vl_2024} & 34.3 \\
    & \nonagentic{} & Qwen2.5-VL (7B)~\citep{bai_qwen25-vl_2025} & 62.4 \\
    & \nonagentic{} & Qwen3-VL (4B)~\citep{bai_qwen3-vl_2025} & 51.6 \\
    & \nonagentic{} & InternVL2.5 (8B)~\citep{chen_expanding_2025} & 61.5 \\
    & \nonagentic{} & InternVL3 (8B)~\citep{zhu_internvl3_2025} & 62.4\\
    & \nonagentic{} & InternVL3.5 (8B)~\citep{wang_internvl35_2025} & 63.8\\
    & \nonagentic{} & GPT-4o-2024-05-13~\citep{openai_gpt-4o_2024} & 58.5\\
\midrule
\multirow{6}{*}{\rotatebox[origin=c]{90}{Compositional}} 
     & \singleagent{} & IdealGPT~\citep{you_idealgpt_2023} & 41.7 \\
     & \nonagentic{} & ViperGPT~\citep{suris_vipergpt_2023} & 37.9 \\
     & \nonagentic{} & VisRep~\citep{khan_self-training_2024} & 51.4 \\
     & \singleagent{}  & HYDRA~\citep{ke_hydra_2024} & 52.8 \\ \cline{2-4}
     & \multiagent{}  & \methodname{} (Ours) \legendbgp{blue!15}{General} & \textbf{64.9} \\
     & \multiagent{}  & \methodname{} (Ours) \legendbgp{green!20}{Domain-Specific} & 64.7 \\
\bottomrule[0.4mm]
\end{tabular}}
\vspace{2pt}
\end{minipage}
\hfill
\begin{minipage}[t]{0.49\textwidth}
\centering
\captionof{table}{\textbf{Performance on OK-VQA dataset.}}
\label{table:okvqa}
\scalebox{0.8}{
\begin{tabular}{c|c|l|c}
\toprule[0.4mm]
\rowcolor{mygray} \multicolumn{2}{c|}{\textbf{Type}} & \textbf{Method} & \textbf{Acc.}\\
\hline\hline
\multirow{13}{*}{\rotatebox[origin=c]{90}{Monolithic}} 
    & \nonagentic{} & PNP-VQA~\citep{tiong_plug-and-play_2022} & 35.9\\
    & \nonagentic{} & PICa~\citep{yang_empirical_2022} & 43.3 \\
    & \nonagentic{} & BLIP-2~\citep{li_blip-2_2023} & 45.9 \\
    & \nonagentic{} & Flamingo (9B)~\citep{alayrac_flamingo_2022} & 44.7\\
    & \nonagentic{} & MiniGPT-4 (13B)~\citep{zhu_minigpt-4_2023} & 37.5\\
    & \nonagentic{} & LLaVA (13B)~\citep{liu_visual_2023} & 42.5 \\
    & \nonagentic{} & InstructBLIP (13B)~\citep{dai_instructblip_2023} & 47.9 \\
    & \nonagentic{} & Qwen2-VL (7B)~\citep{wang_qwen2-vl_2024} & 28.3 \\
    & \nonagentic{} & Qwen2.5-VL (7B)~\citep{bai_qwen25-vl_2025} & 71.8 \\
    & \nonagentic{} & Qwen3-VL (4B)~\citep{bai_qwen3-vl_2025} & 44.4 \\
    & \nonagentic{} & InternVL2.5 (8B)~\citep{chen_expanding_2025} & 75.2\\
    & \nonagentic{} & InternVL3 (8B)~\citep{zhu_internvl3_2025} & 74.7\\
    & \nonagentic{} & InternVL3.5 (8B)~\citep{wang_internvl35_2025} & 75.7 \\
    & \nonagentic{} & GPT-4o-2024-05-13~\citep{openai_gpt-4o_2024} & 33.4\\
\midrule
\multirow{7}{*}{\rotatebox[origin=c]{90}{Compositional}} 
     & \singleagent{} & IdealGPT~\citep{you_idealgpt_2023} & 19.4\\
     & \nonagentic{} & ViperGPT~\citep{suris_vipergpt_2023} & 40.7 \\
     & \nonagentic{} & VisRep~\citep{khan_self-training_2024} & 46.7 \\
     & \singleagent{}  & HYDRA~\citep{ke_hydra_2024} & 59.4 \\
     & \singleagent{} & DWIM~\citep{ke_dwim_2025} & 62.8 \\ \cline{2-4}
     & \multiagent{}  & \methodname{} (Ours) \legendbgp{blue!15}{General} & 76.0 \\
     & \multiagent{}  & \methodname{} (Ours) \legendbgp{green!20}{Domain-Specific} & \textbf{76.5} \\
\bottomrule[0.4mm]
\end{tabular}}
\vspace{2pt}
\end{minipage}
\small Agentic types: \nonagentic{} non-agentic/non-specified;\; \singleagent{} single-agent;\; \multiagent{} multi-agent.
\end{table}

\begin{table}[t]
\vspace{3mm}
\centering
\caption{\textbf{Quantitative comparison (accuracy) on referring expression comprehension task} on RefCOCO, RefCOCO+, RefCOCOg~\citep{kazemzadeh_referitgame_2014} and Ref-Adv~\citep{akula_words_2020} set. Note there is no training set in Ref-Adv, so all scores are \legendbg{yellow!25}{\revise{domain-transfer}}.}
\scalebox{0.88}{
\begin{tabular}{c|c|l|cccc}
\toprule[0.4mm]
\rowcolor{mygray} \multicolumn{2}{c|}{\textbf{Type}} & \textbf{Method} & \textbf{RefCOCO} & \textbf{RefCOCO+} & \textbf{RefCOCOg} & \textbf{Ref-Adv} \\ \hline \hline
\multirow{11}{*}{\rotatebox[origin=c]{90}{Monolithic}} 
& \nonagentic{} & GLIP-L~\citep{li_grounded_2022} & 55.0 & 51.1 & 54.6 & 55.7 \\
& \nonagentic{} & KOSMOS-2~\citep{peng_kosmos-2_2023} & 57.4 & 50.7 & 61.7 & - \\
& \nonagentic{} & YOLO-World-X~\citep{cheng_yolo-world_2024} & 12.1 & 12.1 & 32.9 & 32.2 \\
& \nonagentic{} & YOLO-World-V2-X~\citep{cheng_yolo-world_2024} & 19.8 & 16.8 & 36.5 & 33.1 \\
& \nonagentic{} & GroundingDINO-T~\citep{liu_grounding_2023} & 61.6 & 59.7 & 60.6 & 60.5 \\
& \nonagentic{} & GroundingDINO-B~\citep{liu_grounding_2023} & 90.8 & 84.6 & 80.3 & 78.0 \\
& \nonagentic{} & SimVG~\citep{dai_simvg_2024} & 94.9 & 91.0 & 88.9 & 74.4 \\
& \nonagentic{} & Florence2-B~\citep{xiao_florence-2_2024} & 94.5 & 91.2 & 88.3 & 72.2 \\
& \nonagentic{} & Florence2-L~\citep{xiao_florence-2_2024} & 95.1 & 92.5 & 90.9 & 71.8 \\
& \nonagentic{} & GPT-4o-2024-05-13~\citep{openai_gpt-4o_2024} & 30.5 & 26.2 & - & - \\
& \nonagentic{} & Qwen2.5-VL-72B~\citep{bai_qwen25-vl_2025} & 94.6 & 92.2 & 90.3 & - \\ \hline

\multirow{8}{*}{\rotatebox[origin=c]{90}{Compositional}} 
& \nonagentic{} & Code-bison~\citep{stanic_towards_2024} & 44.4 & 38.2 & - & -\\
& \nonagentic{} & ViperGPT~\citep{suris_vipergpt_2023} & 68.6 & 73.8 & 68.7 & 58.2 \\
& \nonagentic{} & VisRep~\citep{khan_self-training_2024} & 55.2 & 51.1 & - & - \\
& \singleagent{} & HYDRA~\citep{ke_hydra_2024} & 65.7 & 66.2 & 59.9 & 48.3 \\
& \singleagent{} & DWIM~\citep{ke_dwim_2025} & 82.7 & 74.2 & - & - \\
& \singleagent{} & NAVER~\citep{cai_naver_2025} & 96.2 & 92.8 & 91.6 & 75.4 \\ \cline{2-7}
& \multiagent{} & \methodname{} (Ours) \legendbgp{blue!15}{General} & 96.3 & 93.8 & 90.7 & \textbf{77.3} \\
& \multiagent{} & \methodname{} (Ours) \legendbgp{green!20}{Domain-Specific} & \textbf{96.3} & \textbf{93.9} & \textbf{90.8} & - \\
\bottomrule[0.4mm]
\end{tabular}}
\label{tab:detection}
\small Agentic types: \nonagentic{} non-agentic/non-specified;\; \singleagent{} single-agent;\; \multiagent{} multi-agent.
\end{table}

\subsection{Quantitative Results}

\paragraph{Compositional Image Question Answering.} On GQA~\citep{hudson_gqa_2019}, which emphasizes complex compositional reasoning over spatial relations and attributes, \methodname{} reaches 64.9\% accuracy (\autoref{table:gqa}), surpassing previous trainable compositional methods HYDRA and VisRep, training-free baselines such as ViperGPT. It is also competitive with strong monolithic VLMs, exceeding InternVL3.5 and Qwen2.5-VL. The gains stem from the learned transition policy, and the hyper agent understands the capacity of agents. Easy queries invoke \textsc{Specialized} perception first and escalate to \textsc{Oneshot} or \textsc{Stepwise} only on failure or low confidence, whereas difficult cases route directly to \textsc{Stepwise} to maximize the reasoning. When the range of data is narrow and distinctive, the \legendbg{green!20}{domain-specific} setting can calibrate priors more precisely; when compositional patterns are shared across sources, joint training \legendbgp{blue!15}{general} regularizes transitions and reduces overfitting. In GQA we observe the latter, many patterns appear across sources in \datasetname{}, so the \legendbg{blue!15}{general} setting achieves better performance.

\paragraph{External Knowledge-Dependent Image Question Answering.} On OK-VQA~\citep{marino_ok-vqa_2019}, which requires external knowledge, \methodname{} achieves \textbf{76.5\%} accuracy (\autoref{table:okvqa}), surpassing prior compositional systems such as DWIM (62.8\%) and HYDRA (59.4\%), respectively, and outperforming recent monolithic VLMs including Qwen2.5-VL (71.8\%) and InternVL3.5 (75.7\%). Gains come from the learned hyper agent transition: for easy queries the hyper agent first invokes \textsc{Specialized} perception and escalates to the \textsc{Stepwise} or \textsc{Oneshot} reasoner only on failure or low confidence; for difficult queries it directly selects \textsc{Stepwise} for multi-step reasoning, with competitive re-entry into \textsc{Specialized} or \textsc{Oneshot} to reason combining the previous findings and new evidence. We observe the \legendbg{green!20}{domain-specific} setting holds a small edge, likely because of the narrow diversity of the reasoning pattern required in the dataset, whereas joint training \legendbgp{blue!15}{general} slightly dilutes these knowledge.

\paragraph{Referring Expression Comprehension.} On popular benchmarks RefCOCO, RefCOCO+, RefCOCOg~\citep{kazemzadeh_referitgame_2014} and Ref-Adv~\citep{akula_words_2020}, \methodname{} obtains state-of-the-art performance (\autoref{tab:detection}). It sets a new state-of-the-art on these datasets, exceeding strong monolithic and compositional baselines. Notably, Ref-Adv only contains a test set, which means the \datasetname{} does not contain the data collected from it, showing promising \revise{domain-transfer} generalizability of \methodname{}. Note that due to learned transition, short simple queries are solved by \textsc{Specialized} perception with verification, while complex cases trigger \textsc{Stepwise} and \textsc{Oneshot} reasoning. \legendbg{green!20}{Domain-specific} SFT is slightly stronger because the language query styles is dataset-specific.

\begin{table}[t]
\centering
\caption{\textbf{Ablation of hyper agent.} In this table, we report the accuracy for all VQA and referring expression comprehension benchmarks, \revise{and the inference time per query (tested on GQA). \emph{HA: Hyper Automaton. Transition: Transition policy ($\delta_\theta$). SFT: Supervised finetuning.}} Refer to \autoref{sec:ablation_studies} for details.}
\label{table:ablation_hyper_agent}\scalebox{0.8}{
{
\begin{tabular}{ccc|cccccc|c}
\toprule[0.4mm]
\rowcolor{mygray} \multicolumn{3}{c|}{\textbf{Components}} & \multicolumn{6}{c|}{\textbf{Accuracy} ($\uparrow$)} & \textbf{Time} ($\downarrow$) \\
\rowcolor{mygray} \textbf{HA} & \textbf{Transition} & \textbf{SFT} & \textbf{GQA} & \textbf{OK-VQA} & \textbf{RefCOCO} & \textbf{RefCOCO+} & \textbf{RefCOCOg} & \textbf{Ref-Adv} & \textbf{Avg Sec.} \\
\hline \hline
\ding{55} & Exhaustive & \ding{55} & 57.7 & 71.5 & 87.7 & 85.6 & 81.7 & 73.1 & 34.58\\
\ding{51} & Random & \ding{55} & 57.1 & 71.1 & 85.3 & 83.8 & 81.1 & 73.2 & 6.91 \\
\ding{51} & LLM & \ding{55} & 58.5 & 75.1 & 95.8 & 93.5 & 88.0 & 76.0 & 8.07\\
\ding{51} & LLM & \ding{51} & \textbf{64.9} & \textbf{76.5} & \textbf{96.3} & \textbf{93.9} & \textbf{90.8} & \textbf{77.3} & \textbf{8.01}\\
\bottomrule[0.4mm]
\end{tabular}}}
\end{table}

\newcommand{\DiagTrainTest}{%
\multirow{2}{*}{%
\begin{tikzpicture}[baseline=(B.base)]
  \node[inner sep=0pt,outer sep=0pt,
        minimum width=2.5cm,
        minimum height=2.2em,
        fill=mygray] (B) {};
  \draw (B.north west) -- (B.south east); 
  \node[anchor=north east, xshift=-0.35em, yshift=-0.15em, font=\bfseries] at (B.north east) {Test};
  \node[anchor=south west, xshift=-0.05em, yshift=-0.20em, font=\bfseries] at (B.south west) {Training};
\end{tikzpicture}}%
}

\begin{table}[t]
\vspace{3mm}
\centering
\caption{\textbf{Generalizability results.} The top-left header cell uses a diagonal split to indicate \emph{Training Data} (rows, $\downarrow$) versus \emph{Test Data} (columns, $\rightarrow$). Diagonal values \legendbgp{green!20}{domain-specific} train and test on the \emph{same} dataset; off-diagonal values evaluate cross-domain/task transfer \legendbgp{yellow!25}{\revise{domain-transfer}}. The last row reports joint training on the whole \datasetname{} dataset \legendbgp{blue!15}{general}. Off-diagonal values are close to the diagonal ones, indicating strong generalizability of the learned transition policy.}
\label{table:generalizability}\scalebox{0.99}{{\begin{tabular}{>{\centering\arraybackslash}p{2.3cm}||cc|cccc}
\toprule[0.4mm]
\DiagTrainTest & \multicolumn{2}{c|}{\cellcolor{mygray}\textbf{VQA}} & \multicolumn{4}{c}{\cellcolor{mygray}\textbf{Visual Grounding}}\\
& \cellcolor{mygray}\textbf{GQA} & \cellcolor{mygray}\textbf{OK-VQA} & \cellcolor{mygray}\textbf{RefCOCO} & \cellcolor{mygray}\textbf{RefCOCO+} & \cellcolor{mygray}\textbf{RefCOCOg} & \cellcolor{mygray}\textbf{Ref-Adv} \\
\hline \hline
GQA     & \cellcolor{green!20} 64.7 & \cellcolor{yellow!25} 75.8 & \cellcolor{yellow!25} 96.1 & \cellcolor{yellow!25} 93.7 & \cellcolor{yellow!25} 90.4 & \cellcolor{yellow!25} 77.0 \\
OK-VQA  & \cellcolor{yellow!25} 64.1 & \cellcolor{green!20} 76.5 & \cellcolor{yellow!25} 96.2 & \cellcolor{yellow!25} 93.8 & \cellcolor{yellow!25} 90.5 & \cellcolor{yellow!25} 76.9 \\ \hline
RefCOCO  & \cellcolor{yellow!25} 63.8 & \cellcolor{yellow!25} 75.5 & \cellcolor{green!20} 96.3 & \cellcolor{yellow!25} 93.9 & \cellcolor{yellow!25} 90.8 & \cellcolor{yellow!25} 77.2 \\
RefCOCO+  & \cellcolor{yellow!25} 63.6 & \cellcolor{yellow!25} 75.4 & \cellcolor{yellow!25} 96.2 & \cellcolor{green!20} 93.9 & \cellcolor{yellow!25} 90.7 & \cellcolor{yellow!25} 77.1 \\
RefCOCOg & \cellcolor{yellow!25} 63.1 & \cellcolor{yellow!25} 75.4 & \cellcolor{yellow!25} 96.1 & \cellcolor{yellow!25} 93.7 & \cellcolor{green!20} 90.8 & \cellcolor{yellow!25} 77.2 \\ \hline
All     & \cellcolor{blue!15} 64.9 & \cellcolor{blue!15} 76.0 & \cellcolor{blue!15} 96.3 & \cellcolor{blue!15} 93.8 & \cellcolor{blue!15} 90.7 & \cellcolor{blue!15} 77.3 \\
\bottomrule[0.4mm]
\end{tabular}}}
\end{table}

\begin{figure}[t]
\centering
\begin{minipage}[t]{0.64\textwidth}
\includegraphics[width=\textwidth]{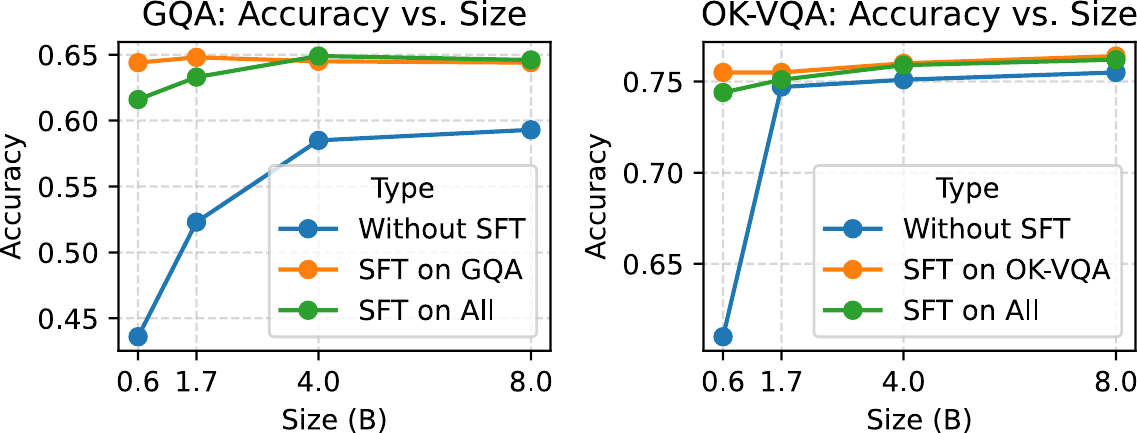}
\caption{\textbf{Results of different LLM sizes.} Accuracy versus the model size (in billions of parameters) of the hyper agent's LLM state controller. Left: GQA; right: OK-VQA. X-axis: LLM size; Y-axis: accuracy.}
\label{fig:llm_size}
\end{minipage}\hfill
\begin{minipage}[t]{0.31\textwidth}
\includegraphics[width=\textwidth]{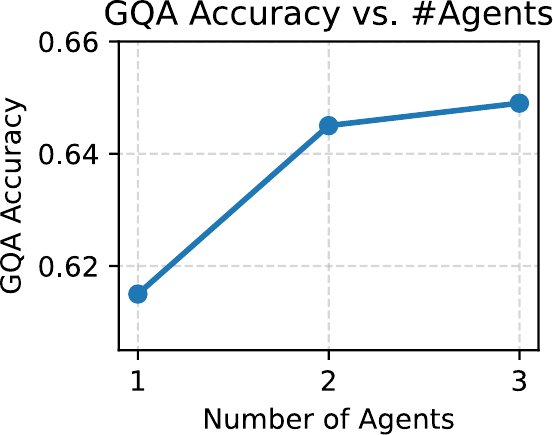}
\caption{\revise{\textbf{Results of different numbers of sub-agents.} X-axis: number of sub-agents; Y-axis: accuracy in GQA.}}
\label{fig:num_agents}
\end{minipage}
\end{figure}

\subsection{Ablation Studies}
\label{sec:ablation_studies}

\paragraph{Hyper Agent.} \autoref{table:ablation_hyper_agent} isolates the main contribution of the trainable hyper agent \revise{and the hierarchical automaton design. We compare: (1) \textbf{Exhaustive Ensemble} without hierarchical automaton (HA): exhaustively call all sub-agents and aggregate with a VLM; (2) \textbf{Random Transition}: HA enabled but the next state is chosen randomly; (3) \textbf{LLM without SFT:} a pretrained LLM is used as the state controller (no supervised finetuning); (4) \textbf{LLM + SFT:} a supervised finetuned LLM controls transitions. Both the exhaustive baseline and random transition yield the weakest performance, but introducing the hyper automaton already cuts runtime significantly.} Replacing random with a pretrained LLM in hyper agent improves accuracy across tasks. This suggests that \revise{(i) the hyper automaton and the LLM primarily drive effective multi-agent collaboration and competition} and (ii) SFT further helps the understanding of the capacity of agents in different types of questions.

\paragraph{Generalizability.} \label{para:generalizability} We conduct generalization analysis by training the hyper agent on GQA subset only of \datasetname{} dataset, OK-VQA subset only, or the whole dataset. \autoref{table:generalizability} organizes results by different training/evaluation types: \legendbg{green!20}{domain-specific}, \legendbg{yellow!25}{\revise{domain-transfer}}, and \legendbg{blue!15}{general}. \revise{domain-transfer} performance is strong in both directions (GQA$\to$OK-VQA; OK-VQA$\to$GQA) with less than 1\% difference. The model trained on all data reaches similar performance to the model trained on the corresponding subset only, indicating the controller learns a task-agnostic transition policy with minimal negative impact. We further discuss the effects in the next paragraph.

\paragraph{LLM Size.} \autoref{fig:llm_size} compares the sizes of the LLM state controller from 0.6B to 8B under three settings: (i) no SFT, (ii) domain-specific SFT, and (iii) SFT on all. With domain-specific SFT, even small models (0.6B/1.7B) perform competitively matching 4B and 8B. When finetuned jointly on all data, small models are worse than 4B/8B by a few percentage points, indicating limited capacity to absorb cross-task knowledge. Without SFT, accuracy drops sharply for smaller models and improves mainly with size. Balancing accuracy and efficiency, we choose 4B as default, as it produces near-optimal results with substantially lower memory, while larger models yielding only marginal gains.

\revise{\paragraph{Number of Agents.} We ablate the number of agent states to quantify benefits beyond our 3-agent design. On GQA, a single \emph{Specialized} agent reaches 61.5\%, adding the \emph{Oneshot} reasoner lifts accuracy to 64.5\%, and adding the \emph{Stepwise} reasoner yields a marginal further gain to 64.9\% (\autoref{fig:num_agents}). The small improvement from 2 to 3 agents indicates diminishing improvements on current benchmarks, suggesting that the agent count is not the major factor. We therefore use three agents in \methodname{}.}

\revise{\paragraph{More Analysis.} We discuss more analysis for generalizability in \autoref{sec:more_analysis_generalizability}, hyper agent in \autoref{sec:more_analysis_hyper_agent}, efficiency in \autoref{sec:more_analysis_efficiency}, comparison with direct SFT in \autoref{sec:more_analysis_compare_direct_sft}, and the qualitative examples in \autoref{sec:qual_analysis} in supplementary materials.}

\section{Conclusion}

We present \methodname{}, a visual reasoning method that uses a trainable hyper agent to learn the transition policy of a hierarchical finite-state automaton. By transitioning between agents based on a shared memory, the system reduces hallucinations, and preserves explainability through explicit states and context. To supervise the hyper agent, we introduced the transition-trajectory dataset \datasetname{}, which converts the trajectory data into a standard SFT format and adapts as agents are added. From experiments, \methodname{} achieves state-of-the-art performance across multiple datasets. \textbf{Limitations.} The data generation pipeline performs a near-exhaustive transition search over the state space; this is tractable with the current three agents but may become costly as the number of states grows.

\subsubsection*{Acknowledgments}
This research is sponsored by the DARPA Assured Neuro Symbolic Learning and Reasoning (ANSR) program under award number FA8750-23-2-1016.

\bibliography{references}
\bibliographystyle{iclr2026_conference}

\clearpage

\appendix

\section{The Use of Large Language Models}
We declare that LLMs (GPT-5/5.1/5.2) are used for the paper language polishing.

\begin{figure}[b]
\centering
\includegraphics[width=\textwidth]{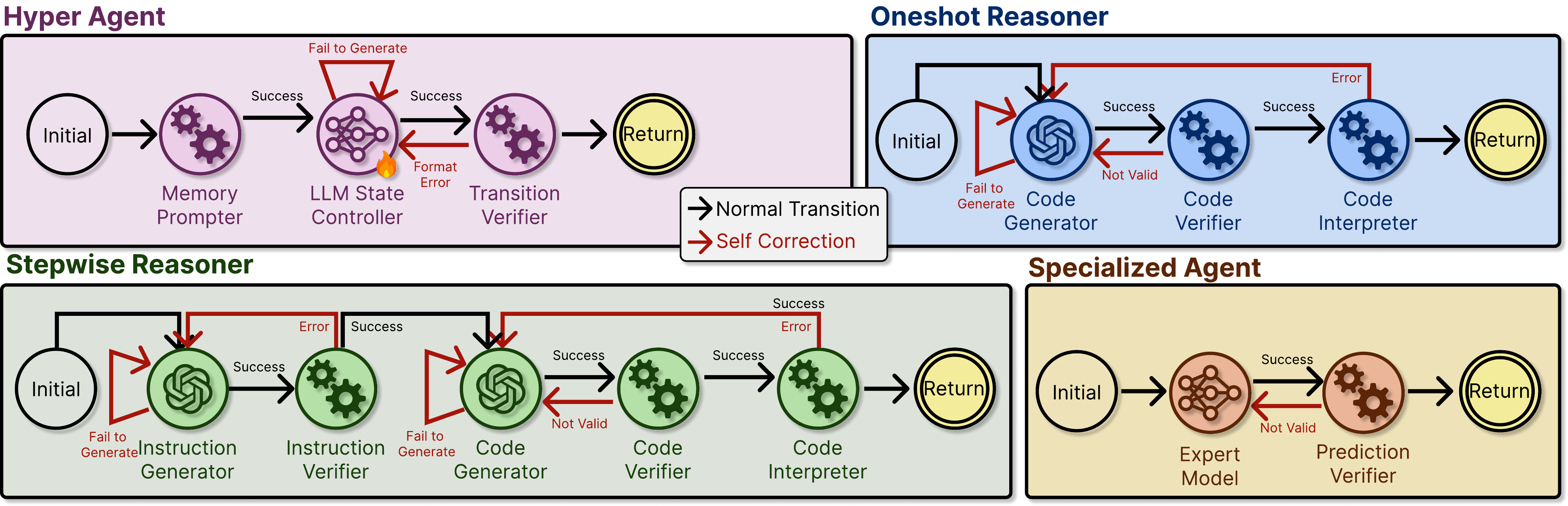}
\caption{\textbf{Details of agents in \methodname{}.} Each block shows the sub-automaton executed when the hyper automaton transits into that agent. Black arrows indicate the normal paths; red arrows show local error-correction paths. Persistent failures transition to \textsc{Failure} state of the hyper automaton (omitted for clarity).}
\label{fig:agents}
\end{figure}

\section{Implementation of Agents}
\label{sec:implementation_agents}

In this section we introduce the detailed implementation of the four agents shown in \autoref{fig:agents}. The hyper agent is triggered at each decision point of the hyper automaton (on \textsc{Initial}, and after any agent returns) then summarizes the shared memory $m_t$ and applies the learned transition $\delta_\theta$ to select the next state $s_{t+1}$. Other agents are triggered only when selected by the hyper agent. Upon entry, the selected agent always starts at its internal \textsc{Initial} state; reaching the agent’s \textsc{Return} state hands control back to the hyper automaton.

\revise{We implemented \emph{three} agents to span levels of reasoning: a \emph{Specialized} System-1 perception agent, an \emph{Oneshot} fast thinking agent, and a \emph{Stepwise} slow thinking agent. Each agent brings different trade-offs. The \emph{Specialized} agent is fast and verifiable for easier subtasks such as finding an object without complex relations, but lacks depth for multi-step compositional reasoning. The \emph{Oneshot} reasoner is cheap and effective on moderately compositional queries, yet might fail on edge cases because it generates the full workflow without accessing the intermediate variable in the workflow. The \emph{Stepwise} agent is designed for complex reasoning via verified program execution, but incurs higher latency and cost. The formulation is modular and scales to additional agents without changing the other part of the system.}

\subsection{Hyper Agent}
As illustrated in \autoref{fig:agents} (top-left), the hyper agent is triggered from hyper automaton, uses the \emph{Memory Prompter} to convert the current shared memory snapshot $m_t$ into a text prompt $x_t$, and feeds it to the trainable \emph{LLM State Controller} to propose the next state. If the LLM fails to generate a valid proposal, we re-prompt once with extra feedback. The output is then checked by the \emph{Transition Verifier}, which enforces valid state selection. On success, the hyper agent returns the chosen state to the hyper automaton and appends the decision to $m_t$.

\subsection{Oneshot Reasoner}
In \autoref{fig:agents} (top-right), the oneshot reasoner enters at \emph{Initial} from the hyper automaton, calls the \emph{Code Generator} to produce a Python program, and passes it to the \emph{Code Verifier} for format checks; generation or validation failure triggers a regeneration. Verified code is executed by the \emph{Code Interpreter} in a Python environment; runtime errors trigger regeneration with extra feedback. On success, the program, execution history, and feedback are appended to $m_t$ and the agent returns to the hyper automaton; if verification or execution repeatably fails, the agent triggers \textsc{Failure}, and returns to the hyper automaton.

\subsection{Stepwise Reasoner}
The stepwise reasoner (\autoref{fig:agents}, bottom-left) handles more complex, slower reasoning: from hyper automaton (\emph{Initial}), the \emph{Instruction Generator} proposes the next one-step plan based on $m_t$, which the \emph{Instruction Verifier} validates; a verification error or failure to generate triggers one regeneration. The accepted plan is translated by the \emph{Code Generator}, checked by the \emph{Code Verifier}, and executed by the \emph{Code Interpreter}; each stage includes error-correction loops as annotated in the figure. If execution succeeds, new context (variables, history, feedback) is written into $m_t$ and the agent returns to the hyper automaton; if any stage stays invalid after multiple attempts, the agent triggers \textsc{Failure} and returns control to the hyper automaton.

\subsection{Specialized Agent}
As shown in \autoref{fig:agents} (bottom-right), the specialized agent begins from hyper automaton, runs an \emph{Expert Model} (\eg, VLM, object detector), and its output is verified by the \emph{Prediction Verifier} for extra checks. If the output is not valid, the agent performs one adaptive retry; otherwise it commits the intermediate results and verifier feedback to $m_t$ and returns to the hyper automaton. Persistent invalid results trigger failure and return control to the hyper automaton.

\newcommand{\DiagTrainTestS}{%
\multirow{2}{*}{%
\begin{tikzpicture}[baseline=(B.base)]
  \node[inner sep=0pt,outer sep=0pt,
        minimum width=3.5cm,
        minimum height=2.2em,
        fill=mygray] (B) {};
  \draw (B.north west) -- (B.south east); 
  \node[anchor=north east, xshift=-0.35em, yshift=-0.15em, font=\bfseries] at (B.north east) {Test};
  \node[anchor=south west, xshift=-0.05em, yshift=-0.20em, font=\bfseries] at (B.south west) {Training};
\end{tikzpicture}}%
}

\begin{table}[tb]
\centering
\caption{\textbf{More generalizability results.} The top-left header cell uses a diagonal split to indicate \emph{Training Data} (rows, $\downarrow$) versus \emph{Test Data} (columns, $\rightarrow$). 
Row \emph{Single the same dataset} trains each LLM state controller in hyper agent on each training set of the dataset and tests on the test set of the same dataset \legendbgp{green!20}{domain-specific}; 
row \emph{All exclude the dataset} trains on the union of the remaining datasets and tests on the held-out column dataset \legendbgp{yellow!25}{\revise{domain-transfer}}; 
row \emph{All include the dataset} trains jointly on all datasets \legendbgp{blue!15}{general}. 
Off-domain accuracies are close to the domain-specific ones, indicating that the learned transition policy generalizes across tasks.}

\label{table:more_generalizability}\scalebox{1}{
{
\begin{tabular}{>{\centering\arraybackslash}p{3.3cm}||cc|ccc}
\toprule[0.4mm]
\DiagTrainTestS & \multicolumn{2}{c|}{\cellcolor{mygray}\textbf{VQA}} & \multicolumn{3}{c}{\cellcolor{mygray}\textbf{Visual Grounding}}\\
& \cellcolor{mygray}\textbf{GQA} & \cellcolor{mygray}\textbf{OK-VQA} & \cellcolor{mygray}\textbf{RefCOCO} & \cellcolor{mygray}\textbf{RefCOCO+} & \cellcolor{mygray}\textbf{RefCOCOg}\\
\hline \hline
Single the same dataset & \cellcolor{green!20} 64.7 & \cellcolor{green!20} 76.5 & \cellcolor{green!20} 96.3 & \cellcolor{green!20} 93.9 & \cellcolor{green!20} 90.8 \\
All exclude the dataset & \cellcolor{yellow!25} 63.5 & \cellcolor{yellow!25} 75.4 & \cellcolor{yellow!25} 96.1 & \cellcolor{yellow!25} 93.8& \cellcolor{yellow!25} 90.7 \\
All include the dataset & \cellcolor{blue!15} 64.9 & \cellcolor{blue!15} 76.0 & \cellcolor{blue!15} 96.3 & \cellcolor{blue!15} 93.8 & \cellcolor{blue!15} 90.7 \\
\bottomrule[0.4mm]
\end{tabular}}}
\end{table}

\section{More Analysis for Generalizability}
\label{sec:more_analysis_generalizability}

We conducted further generalization analysis by training the hyper agent with several more dataset configurations. As shown in \autoref{table:more_generalizability}, we classify the results with three different training data configurations: \emph{Single the same dataset} means that the hyper agent is trained on the training set of the test dataset. \emph{All exclude the dataset} means the hyper agent is trained on the whole \datasetname{} dataset but excluding the corresponding training data from the same dataset, to ensure that it is \revise{domain-transfer}. \emph{All include the dataset} means the hyper agent is trained on the whole \datasetname{} dataset which includes the training data from the same dataset to be evaluated. From the extra results, the observation further supports our findings in \autoref{para:generalizability}.

Across all benchmarks, the \legendbg{yellow!25}{\revise{domain-transfer}} setting (\emph{All exclude the dataset}) is around 1 percentage of the \legendbg{green!20}{domain-specific} setting (\emph{Single the same dataset}). The \legendbg{blue!15}{general} model jointly training on the whole dataset (\emph{All include the dataset}) reaches similar performance of \legendbg{green!20}{domain-specific}. The small gaps indicate that the learned transition policy is largely task-agnostic: it transfers across VQA and grounding without per-dataset tuning, and gains from multi-dataset SFT do not harm in-domain accuracy. Practically, this suggests a single hyper agent can be trained once and reused across visual reasoning tasks.

\section{\revise{More Analysis for Hyper Agent}}
\label{sec:more_analysis_hyper_agent}

\revise{We conduct experiments (\autoref{table:more_results_hyper_agent}) using different models as the state controller in hyper agent. We trained the Qwen3-4B (LLM)~\citep{yang_qwen3_2025} and Qwen3-VL 4B (VLM)~\citep{bai_qwen3-vl_2025} on the full set of \datasetname{}. From the results, the state controller is insensitive to the backbone type: swapping the LLM for a VLM yields near-similar performance across all datasets. We therefore adopt the LLM controller to minimize system complexity and resource requirements while retaining performance.}

\begin{table}[h]
\centering
\caption{\revise{\textbf{More results for the state controller model of hyper agent.} All models are trained on the trajectory transitions of the full \datasetname{}.}}
\label{table:more_results_hyper_agent}\scalebox{0.92}{
{
\begin{tabular}{l|cccccc}
\toprule[0.4mm]
\rowcolor{mygray} \textbf{State Controller} & \textbf{GQA} & \textbf{OK-VQA} & \textbf{RefCOCO} & \textbf{RefCOCO+} & \textbf{RefCOCOg} & \textbf{Ref-Adv} \\
\hline \hline
Qwen3 (4B) (LLM) & 64.9 & 76.0 & 96.3 & 93.9 & 90.8 & 77.3 \\
Qwen3-VL (4B) (VLM) & 64.6 & 76.3 & 96.4 & 94.0 & 89.3 & 76.9 \\
\bottomrule[0.4mm]
\end{tabular}}}
\end{table}

\section{\revise{More Analysis for Efficiency}}
\label{sec:more_analysis_efficiency}

\revise{To further analyze the system time and spatial complexity, we collected and calculated the inference time (seconds), LLM API costs (USD, GPT-4o mini) and vRAM usage (GB) per query, between the state-of-the-art monolithic method Qwen2.5-VL (72B)~\citep{bai_qwen25-vl_2025} with 4-bit quantization, the open sourced compositional agentic method HYDRA~\citep{ke_hydra_2024}, the baseline which call all sub-agents exhaustively to aggregate the final answer, and the proposed method \methodname{}. All measurements are taken on a single NVIDIA L40s 48GB GPU on RefCOCO dataset. As shown in \autoref{table:more_results_effi}, \methodname{} attains the best efficiency compared with HYDRA and exhaustive baseline and is comparable to the monolithic baseline, while achieving the lowest API cost and a moderate vRAM usage (substantially below the 72B model).}

\begin{table}[h]
\centering
\caption{\revise{\textbf{More analysis for efficiency.} We compare the inference time in seconds, LLM API costs in USD, and vRAM in GB, on RefCOCO dataset.}}
\label{table:more_results_effi}\scalebox{1}{
{
\begin{tabular}{l|ccc}
\toprule[0.4mm]
\rowcolor{mygray} \textbf{Method} & \textbf{Time (Seconds)} & \textbf{LLM API Cost (USD)} & \textbf{vRAM Usage (GB)}\\
\hline \hline
Qwen2.5-VL (72B) & 6.27 & -- & 43.70 \\
HYDRA & 14.93 & 0.00332 & 17.59 \\
Exhaustive & 32.74 & 0.00531 & 13.08 \\
\methodname{} & 6.10 & 0.00069 & 19.64 \\
\bottomrule[0.4mm]
\end{tabular}}}
\end{table}

\section{\revise{Comparison with Direct SFT}}
\label{sec:more_analysis_compare_direct_sft}

\revise{We compare two paradigms: (i) \emph{direct SFT} of a single VLM (Qwen3-VL (4B)~\citep{bai_qwen3-vl_2025}, InternVL2.5 (8B)~\citep{chen_expanding_2025}) to output answers, and (ii) \emph{\methodname{}}, which finetunes only the hyper agent’s state controller model as a transition policy. As summarized in \autoref{table:direct_sft}, answer-only direct SFT can improve in-domain accuracy but often harms cross-task generalization, which is consistent with catastrophic forgetting of the model’s latent ``think-then-answer'' ability, while \methodname{} maintains strong transfer because it learns transitions between agents rather than a direct monolithic question-to-answer mapping. The direct SFT on the related dataset gains for Qwen3-VL (4B) reflect its lower zero-shot starting point; stronger VLM like InternVL2.5 (8B) is typically harder to improve via answer-only direct SFT. Overall, \methodname{} delivers higher accuracy and more robust cross-task performance than direct SFT.}

\begin{table}[h]
\centering
\caption{\revise{\textbf{Direct SFT vs. \methodname{}.} We compare (i) directly finetuning a VLM baseline (Qwen3-VL 4B, InternVL-2.5 8B) to output answers and (ii) \methodname{} (SFT on hyper agent) on GQA and OK-VQA. All values are accuracy (\%). “--” denotes using public weights without task-specific finetuning. \emph{Training Dataset} indicates which task split was used for SFT. Color codes follow prior tables: \legendbg{green!20}{domain-specific}, \legendbg{yellow!25}{\revise{domain-transfer}}, \legendbg{blue!15}{general}. Note the pretraining data of LLM/VLMs are unknown; colors are for ease of comparison.}}
\label{table:direct_sft}\scalebox{1}{
\begin{tabular}{l|c|cc}
\toprule[0.4mm]
\rowcolor{mygray} \textbf{Method}& \textbf{Training Dataset} & \multicolumn{2}{c}{\textbf{Test Dataset}}\\
\rowcolor{mygray} & & \textbf{GQA} & \textbf{OK-VQA}\\
\hline \hline
Qwen3-VL (4B) & -- & \cellcolor{yellow!25} 51.6 & \cellcolor{yellow!25} 44.4 \\
Qwen3-VL (4B) & GQA & \cellcolor{green!20} 63.2 & \cellcolor{yellow!25} 61.6 \\
Qwen3-VL (4B) & OK-VQA & \cellcolor{yellow!25} 57.5 & \cellcolor{green!20} 68.1 \\
Qwen3-VL (4B) & GQA, OK-VQA & \cellcolor{blue!15} 63.1 & \cellcolor{blue!15} 66.9 \\ \hline
InternVL2.5 (8B) & -- & \cellcolor{yellow!25} 61.5 & \cellcolor{yellow!25} 75.2 \\
InternVL2.5 (8B) & GQA & \cellcolor{green!20} 63.9 & \cellcolor{yellow!25} 64.1 \\
InternVL2.5 (8B) & OK-VQA & \cellcolor{yellow!25} 35.8 & \cellcolor{green!20} 72.4 \\
InternVL2.5 (8B) & GQA, OK-VQA & \cellcolor{blue!15} 63.8 & \cellcolor{blue!15} 65.2 \\ \hline
MATA & -- & \cellcolor{yellow!25} 58.5 & \cellcolor{yellow!25} 75.1\\
MATA & GQA & \cellcolor{green!20} 64.7 & \cellcolor{yellow!25} 75.8\\
MATA & OK-VQA & \cellcolor{yellow!25} 64.1 & \cellcolor{green!20} 76.5\\
MATA & GQA, OK-VQA & \cellcolor{blue!15} 64.9 & \cellcolor{blue!15} 76.4\\
\bottomrule[0.4mm]
\end{tabular}}
\end{table}


\section{Prompt Templates}
\label{sec:prompt_template}

\methodname{} uses LLMs in multiple places, including: (1) A trainable \emph{LLM state controller} in the \emph{Hyper Agent} routes between states by reading a summarized snapshot of the shared memory. (2) An \emph{Instruction Generator} in \emph{Stepwise Reasoner} proposes the next micro-plan. (3) A \emph{Code Generator} in \emph{Stepwise Reasoner} generates Python code for that step. (4) The \emph{Oneshot Reasoner} employs another \emph{Code Generator} to produce a single-pass program. Across roles, prompts are concise, instruction-style templates that expose the relevant slice of shared memory and tool signatures and require outputs in strict JSON/XML blocks for reliable parsing. The prompt template of the \emph{LLM state controller} is shown in \hyperref[prompt:hyper_agent]{prompt 3.1}. The following prompt blocks show detailed prompt templates of the LLMs.

\begin{prompt}[title={Prompt \thetcbcounter: Instruction Generator in Stepwise Reasoner}]
You are an AI assistant designed to assist with compositional visual reasoning tasks providing valid step by step instruction for answering questions and understanding visual information.\\
Instruction Settings\\
--------------------\\
<InstructionSetting>\{instruction\_setting\}</InstructionSetting>\\
Skills Overview\\
---------------\\
The following are the skills that you can use to solve the query:\\
<Skills>\{skills\}</Skills>\\
Task Description\\
----------------\\
Review the task description below to understand the problem context:\\
<TaskDescription>\{task\_title\}\{task\_description\}</TaskDescription>\\
Example Instructions\\
-------------------\\
How to Use these skills:\\
<Examples>\{instruction\_example\}</Examples>\\
User Query\\
----------\\
This is the query you need to solve:\\
<Query>\{query\}</Query>\\
Current Step\\
------------\\
<Step>\{current\_step\}</Step>\\
Previous Instructions\\
---------------------\\
<PreviousInstructions>\{previous\_instructions\}</PreviousInstructions>\\
Previously Executed Code\\
-----------------------\\
<ExecutedCode>\{previous\_code\}</ExecutedCode>\\
Execution Results\\
----------------\\
<ExecutionResults>\{execution\_results\}</ExecutionResults>\\
Available Variables\\
-------------------\\
<Variables>\{variables\_info\}</Variables>\\
-------------------\\
Based on the current context, generate possible next instructions to help solve the query. For each instruction, assign a probability score indicating how promising it will lead to the final answer.\\
Your response must be in this JSON array format:\\
\{"instructions": [
        \{"instruction": "specific instruction", "probability": 0.X\},
        \{"instruction": "another instruction", "probability": 0.Y\},
        ...
]\}
\end{prompt}

\begin{prompt}[title={Prompt \thetcbcounter: Code Generator in Stepwise Reasoner}]
You are a helpful assistant specializing in visual reasoning tasks. Your goal is to generate Python code that solves a visual reasoning query using the provided code API and examples.\\
API Specification\\
-----------------\\
Use the following code API to guide your solution:\\
<CodeAPI>\{code\_api\}</CodeAPI>\\
Task Description\\
----------------\\
Review the task description below to understand the problem context:\\
<TaskDescription>\{task\_title\}\{task\_description\}</TaskDescription>\\
Example Code\\
-----------\\
Here is an example that illustrates the expected format and approach:\\
<Examples>\{code\_example\}</Examples>\\
User Query\\
----------\\
This is the query you need to solve:\\
<Query>\{query\}</Query>\\
Current Step\\
------------\\
<Step>\{current\_step\}</Step>\\
Previous Instructions\\
---------------------\\
<PreviousInstructions>\{previous\_instructions\}</PreviousInstructions>\\
Current Instruction\\
-------------------\\
<Instruction>\{instruction\}</Instruction>\\
Previously Executed Code\\
-----------------------\\
<ExecutedCode>\{previous\_code\}</ExecutedCode>\\
Execution Results\\
----------------\\
<ExecutionResults>\{execution\_results\}</ExecutionResults>\\
Available Variables\\
-------------------\\
<Variables>\{variables\_info\}</Variables>\\
-------------------\\
Generate Python code that solves the query based on the current instruction. Your code should build upon previous steps and use the available variables. Use the code API as shown in the example. Enclose your code in <PythonCode></PythonCode> tags. If your code provides a final answer, assign it to a variable named ``final\_answer''.
\end{prompt}

\begin{prompt}[title={Prompt \thetcbcounter: Code Generator in Oneshot Reasoner}]
You are a helpful assistant specializing in visual reasoning tasks. Your goal is to generate Python code that solves a visual reasoning query using the provided code API and examples.\\
API Specification\\
-----------------\\
Use the following code API to guide your solution:\\
<CodeAPI>\{code\_api\}</CodeAPI>\\
Task Description\\
----------------\\
Review the task description below to understand the problem context:\\
<TaskDescription>\{task\_title\}\{task\_desc\}</TaskDescription>\\
Example for Reference\\
---------------------\\
Here is an example that illustrates the expected format and approach:\\
<Example>\{code\_example\}</Example>\\
User Query\\
----------\\
This is the query you need to solve:\\
<Query>\{query\}</Query>\\
Extra Context\\
----------------\\
<ExtraContext>\{extra\_context\}</ExtraContext>\\
Code Initialization\\
-------------------\\
An instance of the "ImagePatch" class is already provided. Use the following initialization code as the starting point:\\
<ExecutedCode>\\
image\_patch = ImagePatch(image)\\
</ExecutedCode>\\
Instruction:\\
------------\\
Generate Python code that utilizes the provided API and initialization to solve the query enclosed within the <PythonCode></PythonCode> block. Ensure your solution follows the structure and style of the given example. Ensure the variable ``final\_answer'' is assigned to the result of the query.
\end{prompt}

\section{\revise{Dataset Example}}
\label{sec:dataset_example}

\subsection{\revise{Example for VQA}}

\begin{prompt}[title={Example \thetcbcounter: \datasetname{} Example Input}]
<TaskDescription>\\
Compositional image question answering\\
This type of question is intended to return a textual answer to the given question. \\
Please use "final\_answer" as the variable name when providing Python code. Make sure "final\_answer" is string type.\\
E.g., For the question "What sport can you use this for?", please provide the name of the sport as your answer in the final step.\\
E.g., For the question "Is it good weather?", the final answer must be either "yes" or "no".\\
</TaskDescription>\\
<Query>Is the tall clock small or large?</Query>\\
<Instructions></Instructions>\\
<Feedback>\\
Detection result: Only one clock has been detected in original\_image.\\
</Feedback>\\
<Code>\\
image\_patch = ImagePatch(image)\\
\# Find clock in the image\\
clock\_patches = image\_patch.find(["clock"])["clock"]\\
\# Only one clock has been detected\\
clock\_patch = clock\_patches[0]\\
</Code>\\
<Variables>\\
image\_patch: ImagePatch(0, 0, 500, 333), patch name: original\_image\\
clock\_patches: [ImagePatch(234, 131, 285, 182)]\\
clock\_patch: ImagePatch(234, 131, 285, 182), patch name: clock\_1\_in\_original\_image\\
</Variables>\\
<StateHistory>\\
Initial\\
StepWiseReasoning\\
StepWiseReasoning\\
</StateHistory>\\
<State>StepWiseReasoning</State>\\
<CurrentStep>3</CurrentStep>\\
Based on the information above, determine the next state the system should transition to. Choose from the following states:\\
<StateCandidates>\\
Final\\
Specialized\\
OneShotReasoning\\
StepWiseReasoning\\
</StateCandidates>\\
Return the name wrapped in <NextState> tags.
\end{prompt}

\begin{prompt}[title={Example \thetcbcounter: \datasetname{} Example Output}]
<NextState>StepWiseReasoning</NextState>
\end{prompt}

\subsection{\revise{Example for grounding}}

\begin{prompt}[title={Example \thetcbcounter: \datasetname{} Example Input}]
<TaskDescription>\\
Referring Expression Comprehension\\
This type of task is to return one image patch in the image that corresponds best to the given query. \\
The object described by the query must exist in the image, and only have one patch. You need to first detect that kind of object in the image and then identify which one matches the description in the query. \\
Please use "final\_answer" as the target image patch name when providing Python code. Make sure only one ImagePatch in "final\_answer".\\
E.g., query is "left woman with shoes," return one of the detected woman patches in the final step, don't return shoes patch.\\
E.g., query is "muffins on the table," return one of the muffin patches in the final step, don't return table patch.\\
E.g., query is "white chaise under window", return one of the chaise patches in the final step, don't return window patch.\\
</TaskDescription>\\
<Query>far right</Query>\\
<Instructions></Instructions>\\
<Feedback>\\
Detection result: 5 people have been detected in original\_image.\\
</Feedback>\\
<Code>\\
image\_patch = ImagePatch(image)\\
\# Find people in the image\\
people\_patches = image\_patch.find(["people"])\\
</Code>\\
<Variables>\\
image\_patch: ImagePatch(0, 0, 640, 427), patch name: original\_image\\
people\_patches: \{"people": [ImagePatch(374, 0, 584, 377), ImagePatch(0, 7, 153, 353), ImagePatch(200, 47, 361, 408), ImagePatch(517, 0, 640, 382), ImagePatch(113, 174, 195, 353)]\}\\
</Variables>\\
<StateHistory>\\
Initial\\
OneShotReasoning\\
StepWiseReasoning\\
</StateHistory>\\
<State>StepWiseReasoning</State>\\
<CurrentStep>3</CurrentStep>\\
Based on the information above, determine the next state the system should transition to. Choose from the following states:\\
<StateCandidates>\\
Final\\
Specialized\\
OneShotReasoning\\
StepWiseReasoning\\
</StateCandidates>\\
Return the name wrapped in <NextState> tags.
\end{prompt}

\begin{prompt}[title={Example \thetcbcounter: \datasetname{} Example Output}]
<NextState>StepWiseReasoning</NextState>
\end{prompt}

\section{\revise{Qualitative Analysis}}\label{sec:qual_analysis}

\begin{figure}[h]
\centering
\captionsetup[sub]{font=small}

\vspace{8pt}
\centering
\begin{tcolorbox}[colback=gray!3, colframe=black!10, boxrule=0.4pt, left=6pt, right=6pt, top=6pt, bottom=6pt]
\begin{minipage}{0.27\linewidth}
\centering
\includegraphics[width=\linewidth]{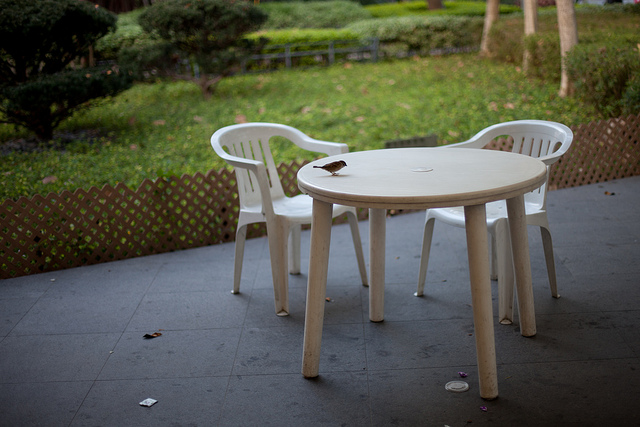}\\[-2pt]
{\scriptsize \textit{Example from GQA}}
\end{minipage}
\hfill
\begin{minipage}{0.70\linewidth}
\scriptsize
\textbf{Query:} \emph{``What's in front of the fence?"} \textbf{Ground truth:} \emph{chair} \\[2pt]
\textbf{Qwen3-VL (4B):} A \textcolor{red}{small bird} perched on the table. $\rightarrow$ \ding{55} incorrect.\\
\textbf{ViperGPT:} Generates a single program; mis-detects objects \textcolor{red}{``table''} $\rightarrow$ \ding{55} incorrect.\\
\textbf{HYDRA:} Multistep programming reasoning, but relies on the noisy tools output and unable to fix, producing \textcolor{red}{``There are no objects in front of the fence.''} $\rightarrow$ \ding{55} incorrect.\\
\textbf{NAVER:} Follows fixed Perception$\rightarrow$Logic$\rightarrow$Answering; \textcolor{red}{cannot produce the text answer} $\rightarrow$ \ding{55} incorrect.\\
\textbf{\methodname{} (ours):} Initially the hyper agent calls \underline{Stepwise Reasoner} 3 steps, with detailed exploration and trials, but failed to get the result. Then inherited with the experience of the memory, the hyper agent decides to transit to \underline{Oneshot Reasoner}, which generates a correct answer based on the previous experience: ``white \textcolor{ForestGreen}{chair}'' $\rightarrow$ \ding{51} correct. 
\end{minipage}
\end{tcolorbox}
\begin{tcolorbox}[colback=gray!3, colframe=black!10, boxrule=0.4pt, left=6pt, right=6pt, top=6pt, bottom=6pt]
\begin{minipage}{0.27\linewidth}
\centering
\includegraphics[width=\linewidth]{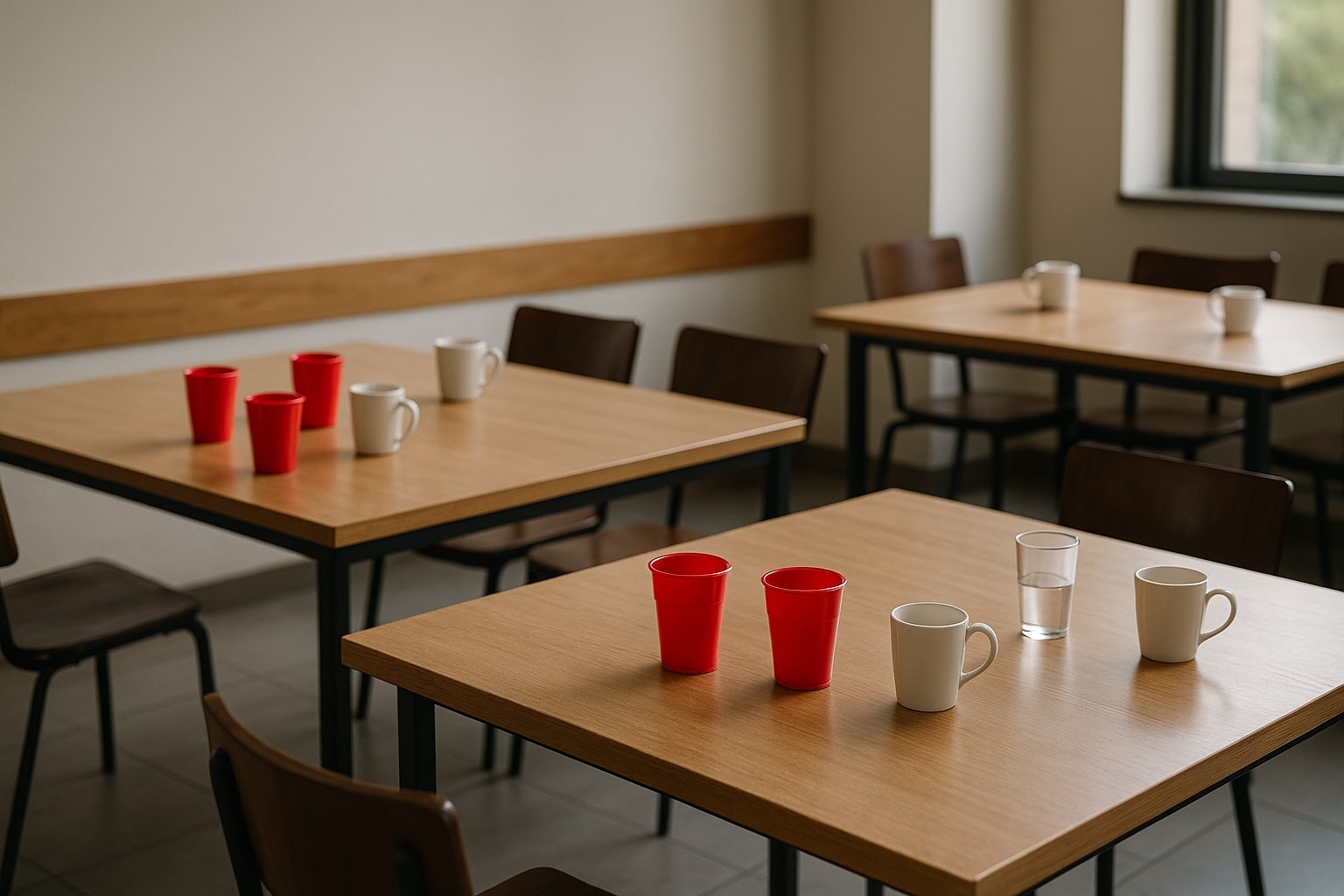}\\[-2pt]
{\scriptsize \textit{Example zero-shot}}
\end{minipage}
\hfill
\begin{minipage}{0.70\linewidth}
\scriptsize
\textbf{Query:} \emph{``How many red cups on the left table?"} \textbf{Ground truth:} \emph{3}\\[2pt]
\textbf{Qwen3-VL (4B):} ... (thinking) There are \textcolor{red}{4} red cups on the left table. $\rightarrow$ \ding{55} incorrect.\\
\textbf{ViperGPT:} Generates a faulty program; no mechanism to fix the program, \textcolor{red}{cannot produce answer} $\rightarrow$ \ding{55} incorrect.\\
\textbf{HYDRA:} Detects 12 cups and queries color per cup, but never restricts to the left table; answerer then guesses \textcolor{red}{5}  $\rightarrow$ \ding{55} incorrect.\\
\textbf{NAVER:} Manually defined Perception$\rightarrow$Logic$\rightarrow$Answering automaton picks only cup, build the logic that treats ``left table" as left of other cups, then \textcolor{red}{collapses to one localization instead of counting}; with confidence thresholding and counting to the detections it yields the correct count \textcolor{ForestGreen}{3} $\rightarrow$ \ding{51} partial correct. \\
\textbf{\methodname{} (ours):} The hyper agent calls \underline{Oneshot Reasoner} but failed to get a confident answer, then transfers to \underline{Stepwise Reasoner} invoked 4 times to generate the final answer: \textcolor{ForestGreen}{3} $\rightarrow$ \ding{51} correct.
\end{minipage}
\end{tcolorbox}
\caption{\revise{\textbf{Qualitative comparison.} Previous methods either commit to a single pass (ViperGPT), multi-step within one agent (HYDRA), or follow a fixed automaton (NAVER). \textbf{MATA} learns when to \emph{switch agents} and re-enter perception based on shared-memory feedback, yielding robust outcomes on the examples not only from GQA but also the unseen set.}}
\label{fig:qual_comparison}
\end{figure}

\revise{We compare \methodname{} with Qwen3-VL~\citep{bai_qwen3-vl_2025}, ViperGPT~\citep{suris_vipergpt_2023}, HYDRA~\citep{ke_hydra_2024}, and NAVER~\citep{cai_naver_2025}. In easy cases (e.g., ``find people in red''), the \methodname{} hyper agent transits to a \emph{Specialized} agent that answers directly, and most baselines also succeed. For more complex queries (see \autoref{fig:qual_comparison}), stronger compositional reasoning is required; prior methods often hallucinate due to some bottlenecks (e.g., noisy tool outputs, fixed pipelines, no verification).} 

\revise{In \textbf{Example 1 (GQA)}, \methodname{} explores with several \emph{Stepwise Reasoner} steps and, after verification failures, hands off the shared memory to the \emph{Oneshot Reasoner} to understand the previous experience, and produce the correct answer. In \textbf{Example 2} (zero-shot, generated by GPT-Image), it begins with the \emph{Oneshot Reasoner} to build the initial exploration and save to shared memory, then transitions to the \emph{Stepwise Reasoner}, which first isolates the left table and then counts, again yielding the correct result. These cases illustrate how learned transitions improves robustness.}

\end{document}